%% file: FINALE.tex
\documentclass[graybox]{svmult}

\usepackage{type1cm}       

\usepackage{makeidx}         
\usepackage{graphicx}       

\usepackage{multicol}
\usepackage[bottom]{footmisc}

\usepackage{natbib}
\usepackage{newtxtext}
\usepackage[varvw]{newtxmath}

\usepackage{booktabs}
\usepackage{lscape}
\usepackage{longtable}
\usepackage{multirow}
\usepackage{rotating}

\usepackage{soul}
\usepackage{tabularx}
\usepackage{float}

\makeindex
\begin{document}

\title*{A path in regression Random Forest looking for spatial dependence: a taxonomy and a systematic review}
\titlerunning{Spatial dependence in Random Forest}
\author{Luca Patelli, Michela Cameletti, Natalia Golini, Rosaria Ignaccolo}
\institute{Luca Patelli \at University of Pavia, Department of Economics and Management, Via San Felice al Monastero, 5, Pavia, Italy, \email{luca.patelli01@universitadipavia.it}
\and Michela Cameletti \at University of Bergamo, Department of Economics, Via dei Caniana, 2, Bergamo, Italy, \email{michela.cameletti@unibg.it}
\and Natalia Golini \at University of Turin, Department of Economics and Statistics ``Cognetti de Martiis", Lungo Dora Siena, 100A, Torino, Italy, \email{natalia.golini@unito.it}
\and Rosaria Ignaccolo \at University of Turin, Department of Economics and Statistics ``Cognetti de Martiis", Lungo Dora Siena, 100A, Torino, Italy, \email{rosaria.ignaccolo@unito.it}}

\maketitle

\abstract*{}

\abstract{Random Forest (RF) is a well-known data-driven algorithm applied in several fields thanks to its flexibility in modeling the relationship between the response variable and the predictors, also in case of strong non-linearities. In environmental applications, it often occurs that the phenomenon of interest may present spatial and/or temporal dependence that is not taken explicitly into account by RF in its standard version. In this work, we propose a taxonomy to classify strategies according to when (\textit{Pre}-, \textit{In}- and/or \textit{Post-processing}) they try to include the spatial information into regression RF. 
Moreover, we provide a systematic review and classify the most recent strategies adopted to ``adjust'' regression RF to spatially dependent data, based on the criteria provided by the Preferred Reporting Items for Systematic reviews and Meta-Analysis (PRISMA). The latter consists of a reproducible methodology for collecting and processing existing literature on a specified topic from different sources. PRISMA  starts with a query and ends with a set of scientific documents to review: we performed an online query on the 25$^{th}$ October 2022 and, in the end, 32 documents were considered for review. The employed methodological strategies and the application fields considered in the 32 scientific documents are described and discussed. This work falls inside the Agriculture Impact On Italian Air (AgrImOnIA) project.} 

\section{Introduction}
\label{Intro}

In environmental sciences, data are often characterised by spatial and/or temporal information, so that it is possible to identify the spatial location of the observed units and/or to keep track of the same entity over time. Moreover, recent years have been characterised by a huge volume and variety of available data, from several sources and with different resolutions. As a consequence, flexible models are needed in order to describe complex phenomena with available, possibly complex, data.

With regard to the spatial framework, three types of data can be defined: point referenced data, areal data, and point pattern data \citep{cressie1993statistics, banerjee2015hierarchical}.
This work focuses on the first type of data, also known as geostatistical or geocoded data, representing observations of a stochastic process $Y(s)$, where the spatial index $s$ is continuous over a defined region $D\subseteq \mathbb{R}^d$ (usually $d=2$ and $s$ is the \textit{(latitude, longitude)} vector).
The spatial process realisations \(y(s)\) are available only for a specific and limited set of sites $\{s_1,s_2,...,s_n\}$. A classical example of geostatistical data refers to the measurement of air pollutant concentrations obtained through monitoring stations, where the values are observed only in the presence of a monitoring site, although air pollution is distributed continuously in space. 

In the framework of the \textit{Data Modeling Culture} (DMC, \citealp{breiman2001cultures}), among the models available for geostatistical data, 
the gold standard is the Kriging one \citep{cressie1993statistics}, which is a spatial regression model that allows to predict the response variable at unmonitored sites, using the available information from the monitoring network.

However, in general, spatial (parametric) regression models make extensive use of linear algebra operations for the parameters estimation and prediction. Consequently, in the case of a complex model and/or a high number of predictors and/or observations, fitting and prediction could be computationally expensive and, in some cases, even unfeasible (see e.g. \citealp{banerjee2015hierarchical}).

As an alternative to DMC, in recent years, the role of the \textit{Algorithmic Modeling Culture} (AMC) has grown \citep{breiman2001cultures}. In particular, supervised Machine Learning (ML) and Deep Learning (DL) algorithms include non-parametric predictive techniques that do not require assumptions on the relationship between the response variable and the predictors. Being the estimation completely data-driven, they can be extremely flexible and able to model complex non-linear relationships. The counterpart of this flexibility is the missingness of ``interpretability'', if compared with the standard statistical approaches \citep{molnar2022}. Among the algorithms that have been proposed inside the AMC, Random Forest (RF) has been selected for this work because it is widely used in many application fields, it is considered a simple method from an implementation point of view and has a very good predictive performance \citep{balogun2021review}.
Nevertheless, RF is not able to exploit automatically the information coming from the spatial correlation which may exist in the data, and this could have a detrimental effect on the prediction performances. For this reason, it is necessary to adapt standard RF to the spatial framework in order to include in the learning process the spatial information. 

Up to our knowledge, a systematic review of regression RF dealing with spatially correlated data does not exist. 
In literature, interesting review papers have been proposed that discuss the comparison of ML approaches applied to spatial data \citep{nikparvar2021machine} and of ML applications focused on air quality and climate change \citep{balogun2021review}, soil mapping \citep{WADOUX2020review} and raster mapping using GIS \citep{Wylie2019review}.

In order to find a path among the literature about spatial regression RF, the aim of this chapter is twofold. We first propose a new taxonomy, i.e. a scheme of classification specific to scientific documents related to the research topic. The taxonomy aims to classify literature contributions into homogeneous groups based on the adopted strategies, according to when (\textit{Pre}-, \textit{In}- and \textit{Post-processing}) the regression RF algorithm is ``adjusted" to deal with the spatial information. 
Secondly, we present a systematic review of the most recent (from 2010 to 2022) strategies and applications of RF for spatially correlated data. The literature review is performed by applying the Preferred Reporting Items for Systematic reviews and Meta-Analysis (PRISMA) approach, a reproducible methodology for collecting and processing the contributions that are available in the literature at the time of writing \citep{page2021prisma}. In particular, the keywords we used for the PRISMA search originate from the geostatistics terminology and refer to the concept of spatial correlation and spatial dependence.
Our taxonomy is applied here to the output of the PRISMA-based literature review, but it represents a general and consistent classification scheme that can be applied to future reviews and for the proposals of new approaches.

The chapter is structured as follows.
Section \ref{PRISMA} outlines the PRISMA methodology, describing the flow chart diagram for searching and selecting literature contributions.
Section \ref{RFSection} presents standard regression RF and discusses the limitations of using it in the presence of spatial correlation. Section \ref{TaxonomySec} describes the taxonomy we propose to better identify and compare which are in the literature the main contributions of the research topic.
Section \ref{RFStateofArt} presents and discusses the scientific contributions considered in this PRISMA-based literature review. Finally, Section \ref{Conclusions} concludes the chapter with a discussion of the main findings.

%%%%%%%%%%%%%%%%%%%%%%

\section{PRISMA methodology for systematic literature review} \label{PRISMA}
This section explains how the systematic literature review is performed according to the PRISMA methodology \citep{page2021prisma}.
The major scientific documents' sources are online electronic databases, as \textit{Scopus} and \textit{Web of Science} (WoS), which are selected here as primary suppliers.
Aiming to extract records linked to the topic, i.e. RF in the spatial framework, we queried the two databases using the following set of keywords: (``Random Forest" AND (``spatia* dependen*" OR ``spatia* correla*"))\footnote{The keywords are combined using the Boolean Operators, where AND returns a conjunction of the keywords while OR represents a disjunction. The asterisks mean that the written root or a derivative can be used for the query. For example, the use of \textit{spatia*} also includes spatial and spatially. The use of quotes ``..." allows only results in which the exact combination of all words within them is present.}. The first part of the query string refers to the method, while the second considers the spatial framework.
For each database, the query is performed on defined fields of the records. In order to perform an equivalent extraction, the search by keywords is always performed on the contribution title, abstract and keywords. These fields in Scopus are identified using the option \textit{TITLE-ABS-KEY} while in WoS with the option \textit{Topic}. In particular, the latter performs the search also on \textit{Keywords Plus}, a set of words or phrases frequently used in the scientific documents published in WoS.

\begin{figure}[ht]
    \centering
    \includegraphics[width=1\textwidth]{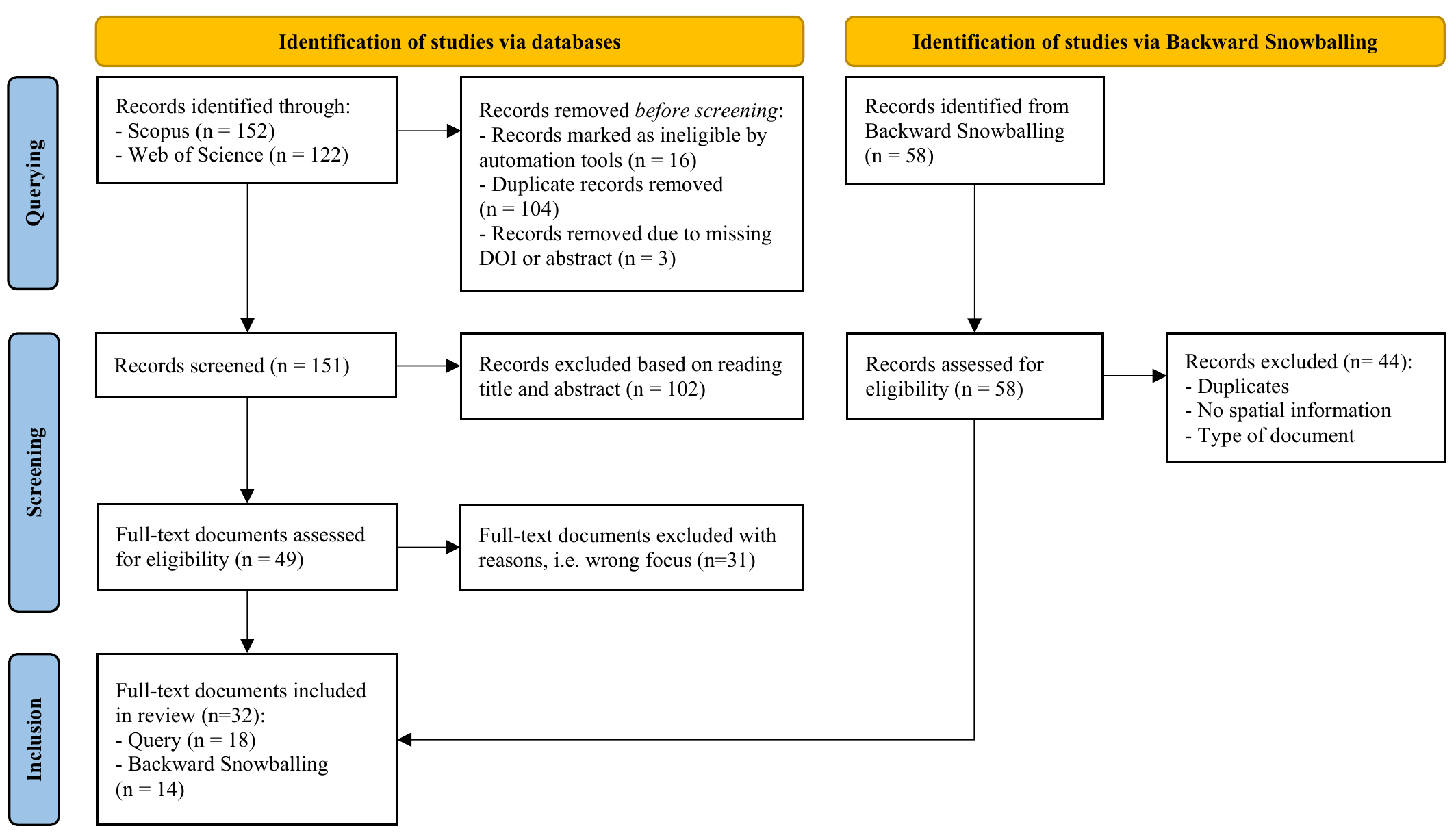}
    \caption{Flow chart diagram for the PRISMA-based systematic review.}
    \label{fig:PRISMA}
\end{figure}

On the query date, \(25^{th}\) October 2022, a total of 274 records were identified. It is important to note that subsequent extractions could result in a different number of records.
Figure \ref{fig:PRISMA} is a graphical representation of the entire literature review process, which consists of querying, screening and including scientific documents related to the specified topic.

In the querying step, we used the following Scopus and WoS automation tools for each database output: 
\begin{itemize}
    \item the considered documents must be Article, Chapter or Conference Paper (the latter was available only for \textit{Scopus});
    \item the source must be a Journal, a Book series, or a Conference proceeding (these were available only for \textit{Scopus});
    \item the documents must be published between 2010 and 2022 (included);
    \item the language must be English.
\end{itemize}
The application of these conditions led to the drop of 16 records. After joining all the remaining 258 records in a dataset, two further selection criteria are applied:
    \begin{itemize}
    \item no duplicates are admitted;
    \item DOI and abstract must be available.
\end{itemize}
As a consequence, 107 records were removed.

At the screening step, 102 out of 151 records were excluded by reading titles and abstracts because neither the title nor the abstract contains hints about the application of RF with some novelties specific to the spatial framework. The resulting 49 full-text documents were assessed for eligibility, and 31 of them were excluded as they had a wrong focus. At this point, we had 18 remaining scientific documents.

Given that the queries performed on the online electronic databases may miss some literature contributions, it was decided to search for further publications using the so-called \emph{Backward Snowballing} \citep{greenhalgh2005effectiveness}.
This procedure consists in retrieving scientific documents by looking for, in the 18 documents already identified, cited references related to the chosen keywords (e.g. spatial correlation). The backward snowballing search led to 58 references. Subsequently, an eligibility assessment was performed by applying the criteria used previously to refine the query's output. In particular, we removed the documents that are duplicates or already appear in the list of the original 18 contributions, have a wrong focus, or do not satisfy the conditions used for the databases search. The resulting 14 documents were appended to the ones already identified via databases, obtaining a set of 32 full-text documents for the systematic review.

%%%%%%%%%%%%%%%%%%%%%%%%%%%%%%%%

\section{Random Forest}
\label{RFSection}
RF is a popular ML algorithm introduced by \cite{breiman2001random} in the context of supervised learning methods, i.e. focused on the prediction of a response variable (output variable) given some predictors (input variables).
RF is considered both for classification and regression problems as it can deal with categorical and quantitative response variables. As an ensemble of decision trees, RF pertains to tree-based methods. This section briefly introduces the basics of decision trees applied to regression problems, standard RF and its limitations in the spatial framework.

\subsection{Regression Trees}
\begin{figure}
    \centering
    \begin{tabular}{cc}
     \includegraphics[width=0.49\textwidth]{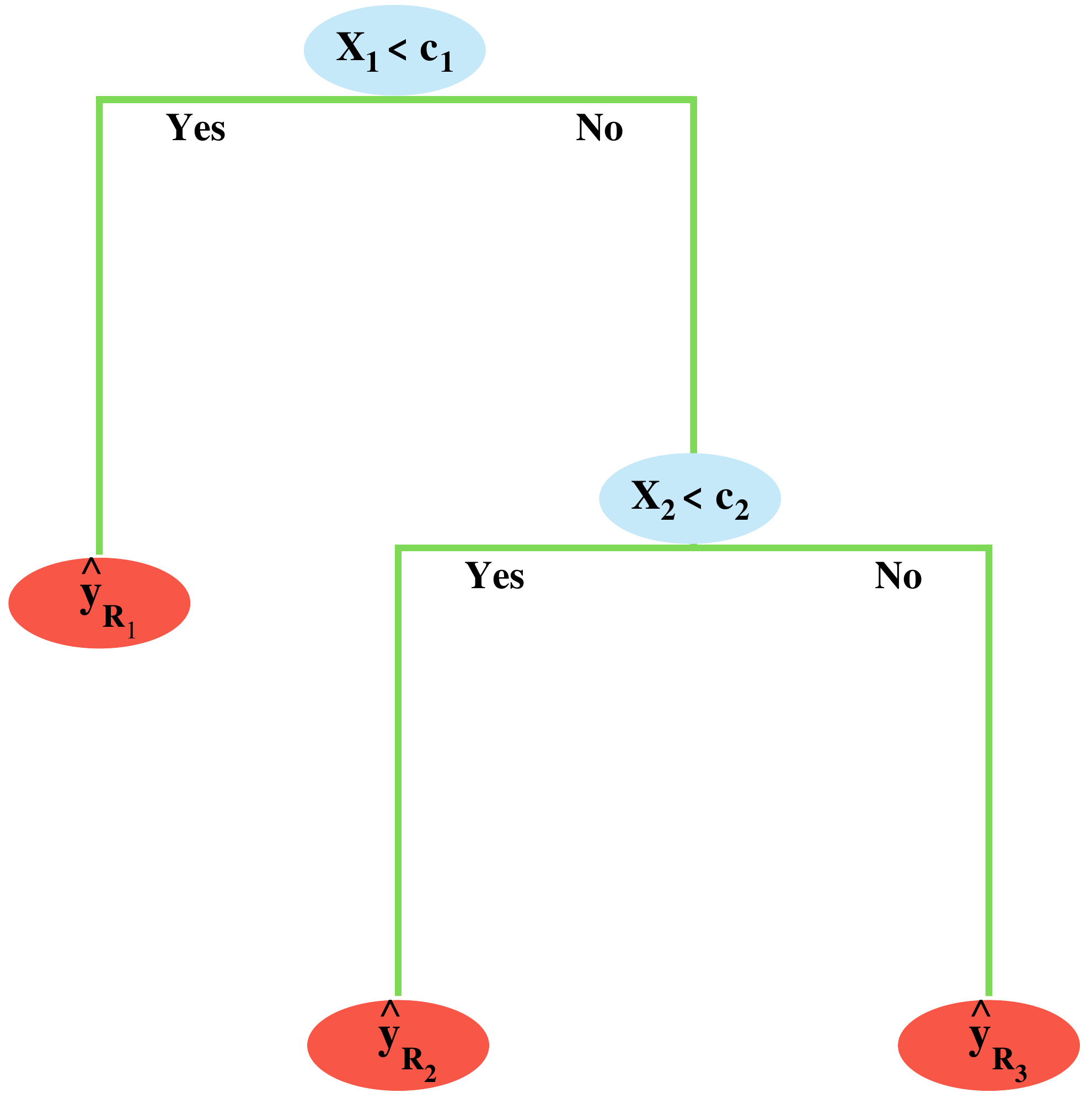} & \includegraphics[width=0.49\textwidth]{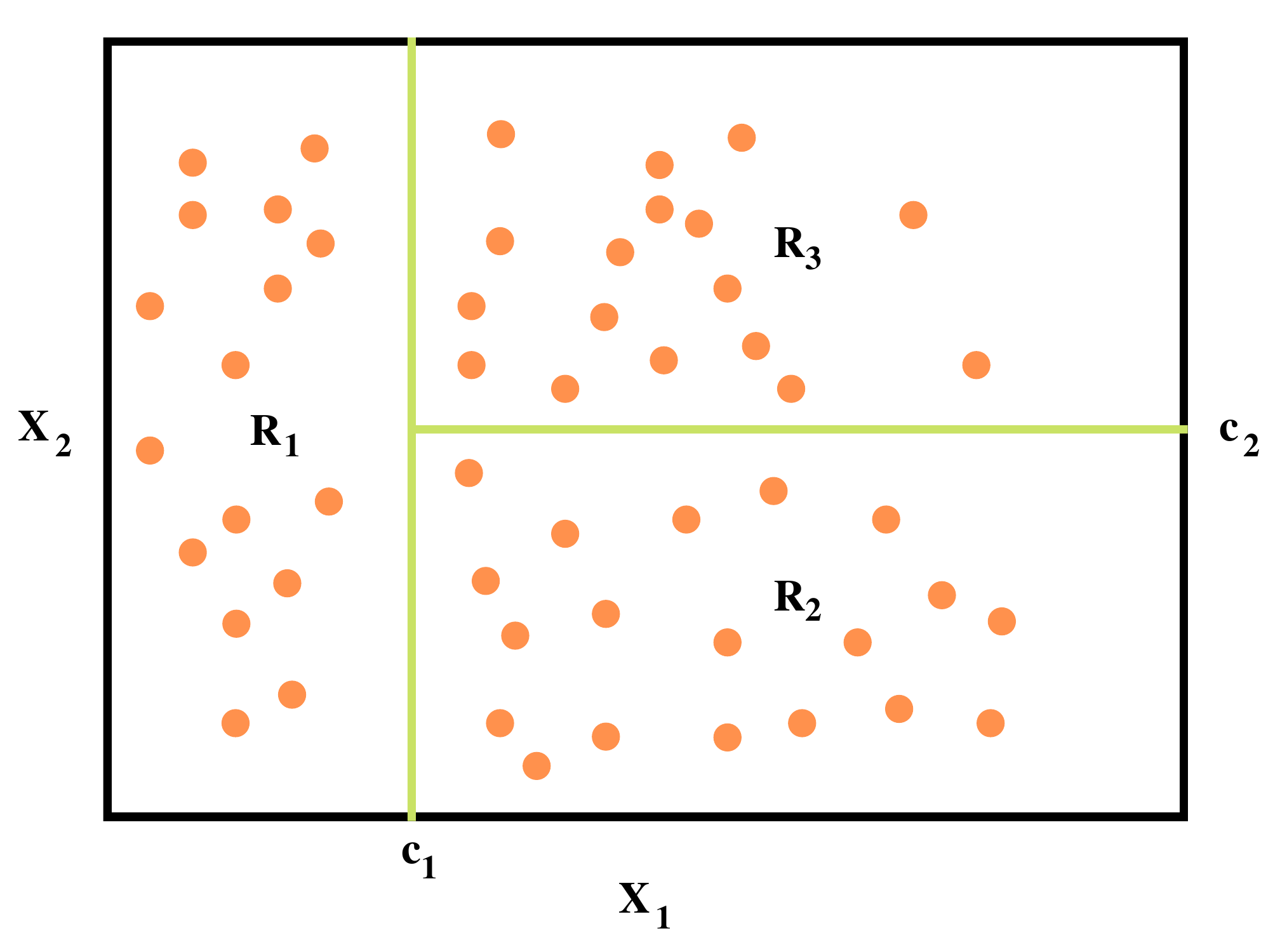} \\
    \end{tabular}
    \caption{Left: illustrative example of a fitted regression tree. Each internal node (in cyan) contains a splitting rule given by the combination of a predictor and a cutpoint value. Each terminal node (in red) shows the response prediction for the (new) observations belonging to the defined region $R_j \ (j=1,2,3)$. The branches (in green) link the nodes and represent the sequential construction of the tree based on a set of logical (Yes/No) tests. Right: the predictor space is partitioned into three regions according to the internal nodes.}
        \label{figDecisionTree}
\end{figure}

A regression tree \citep{breiman1984classification} is a non-parametric learning method which consists of a set of splitting rules used to segment the $P-$dimensional predictor space into $J$ smaller non-overlapping regions (i.e. high-dimensional rectangles). The left panel of Figure \ref{figDecisionTree} shows the graphical representation of a simple regression tree fitted using $P=2$ predictors $X=(X_1, X_2)$. In order to get a prediction for the response variable $Y$, it is necessary to follow the tree from the top (root node) down to a leaf (or terminal node). In the example of Figure \ref{figDecisionTree} (left panel), the root node tests if $X_1$ is lower than the value $c_1$: if this splitting condition is satisfied for a new observation, we move to the left branch which terminates in a leaf node providing the prediction given by $\hat y_{R_1}$. Otherwise, we move to the right branch which leads to the internal node defined by the splitting rule which tests if $X_2$ is lower than the value $c_2$. The tree does not contain other internal nodes and ends with the two terminal nodes on the right: in particular, the value $\hat y_{R_2}$ is the prediction for the (new) observations such that $X_1\geq c_1$ and $X_2<c_2$. The value $\hat y_{R_3}$ instead is the prediction for the observations that fall in the region defined by $X_1\geq c_1$ and $X_2\geq c_2$. The right panel of Figure \ref{figDecisionTree} shows the predictor space segmented into three non-overlapping regions ($R_1, R_2, R_3$) according to the splitting rules and the three terminal nodes illustrated in the left panel.

The process of building a regression tree consists in defining the set of all the binary splitting rules (each specified by a predictor-cutpoint combination). For example, denoting by $X_p$ the generic predictor with $p=1$ or $p=2$, the top node will give rise to the following pair of regions:
\begin{equation}
    R_1(X_p,c_p)=\{X|X_p<c_p\} \quad \text{and} \quad R_2(X_p,c_p)=\{X|X_p\geq c_p\}
\end{equation}
(with respect to the previous example $R_1(X_p,c_p)$ coincides with $R_1$, while $R_2(X_p,c_p)$ with $R_2 \cup R_3$).
The values of $X_p$ and $c_p$ are such that they minimise the Residual Sum of Squares (RSS) given by the sum of two terms, each corresponding to a region:
\begin{equation}
    \sum_{i\in R_1}(y_i-\hat{y}_{R{_1}})^2 \quad + \sum_{i\in R_2}(y_i-\hat{y}_{R{_2}})^2,
\end{equation}
\noindent where $\hat{y}_{R_j}$ ($j=1,2$) is the response variable prediction given by the response variable mean computed using the training observations falling into each region. The binary splitting process is then repeated within each of the resulting two regions, trying to further reduce the RSS or until a stopping rule is satisfied, e.g. a minimum number of training observations into each region or a maximum number of leaves nodes.

This approach is considered greedy because when defining each node, it only takes into account the local error reduction without a global perspective \citep{saha2021random}. It means that the algorithm prefers to reduce the RSS locally at each node, but this may not lead to the best trees, which minimises the global RSS.

Trees are easily flexible and interpretable also thanks to the possibility of representing the relationship between the response variable and predictors graphically, whether quantitative or categorical. Nevertheless, the excess of flexibility, given by the high adaptability to the data, is a limitation of this methodology. In fact, trees are not robust because small changes in the data can introduce large variability in the predictions. This excess of variability is reflected in lower predictive accuracy in comparison to other approaches. It is possible to overcome these limitations by using ensemble methods based on the construction of several trees \citep{james2021introduction}.

\subsection{Standard Random Forest algorithm} \label{standardRFalgorithm}

A first tree ensemble method is represented by bootstrap aggregation, simply known as \textit{bagging}.
Bagging aggregates the predictions from a large number, say $B$, of decision trees grown using $B$ different training data sets created by bootstrapping the training observations \citep{james2021introduction}.
The $B$ trees are usually deep, so their predictions will be characterised by low bias and high variance, whose magnitude is attenuated by aggregating a large number $B$ of such trees.
However, there is one drawback: fully grown trees of bootstrapped data may exhibit a high correlation, especially in the presence of strong predictors, which may appear in the majority of trees. RF deals with this drawback by introducing a novelty: during the splits for the construction of each tree, only a limited random subset of the original predictors (say $m<P$) is taken into account. On average $(P-m)/P$ of the splits will not consider the strong predictors, where $P$ is the total number of predictors and $m$ is the size of the subset of predictors. Usually, for a regression problem $m\approx P/3$. Introducing randomness in the partition phase decorrelates trees leading to results that are less influenced by the most important predictors. Moreover, by reducing the number of times in which the most important predictors are included in the splittings, RF allows identifying the variable influence of all the predictors by computing specific indexes as for example the so-called variable importance \citep{james2021introduction}. 

In the case of a regression problem, the prediction obtained through RF, for a new observation with predictor vector given by \(X\), is the average of the predictions obtained from the \(B\) trees:
\begin{equation}\label{Eq:RFpred}
    \hat{y}_{RF}(X)=\frac{1}{B}\sum_{b=1}^B\hat{y}_{b}(X),
\end{equation}
where $\hat{y}_{b}$ is the prediction given by the \(b\)-th tree grown using a bootstrap sample of training data and only $m$ predictors chosen randomly.

\begin{figure}[ht]
\centering
\includegraphics[width=1\textwidth]{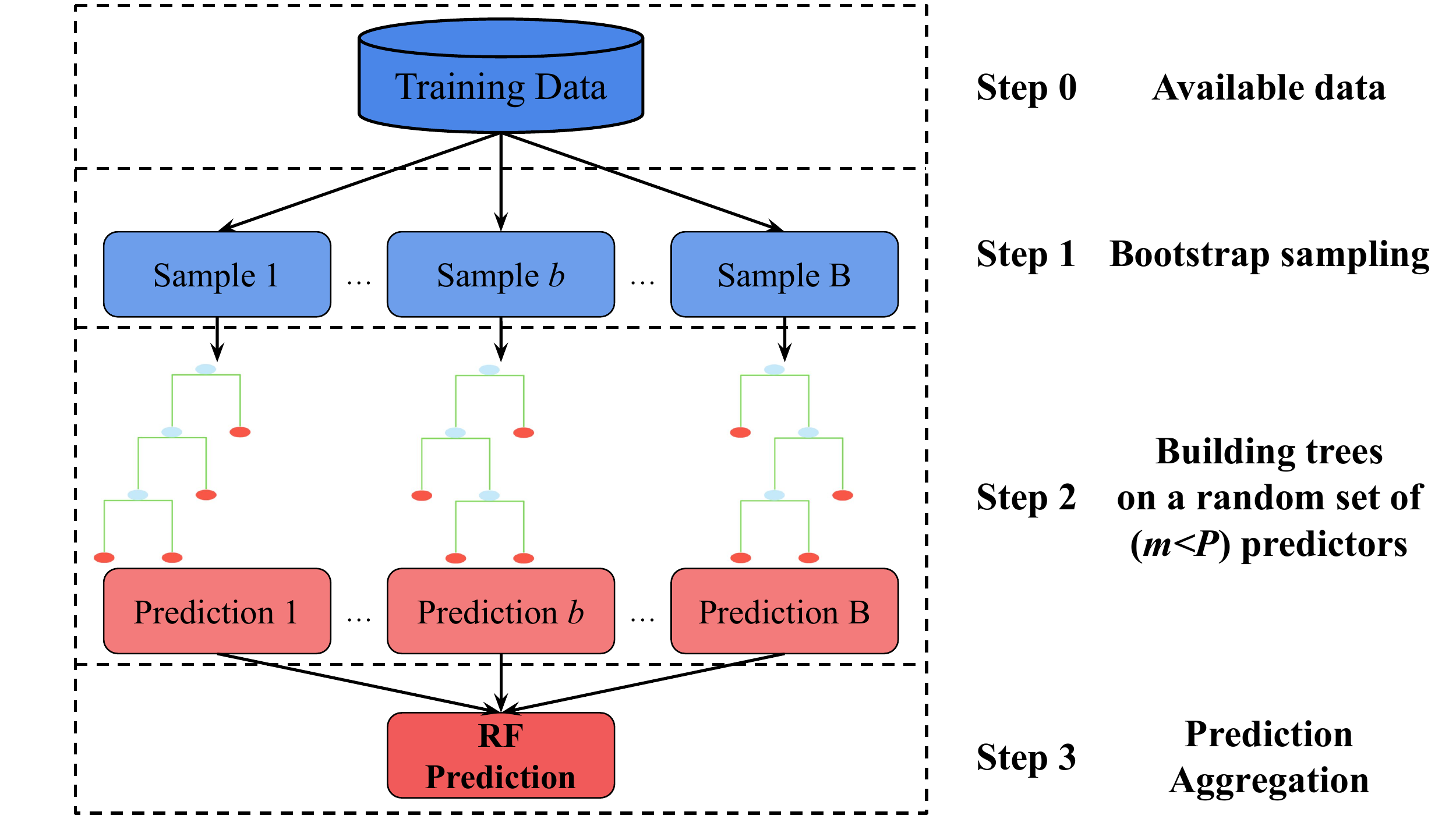}
\caption{Graphical representation of the RF algorithm.}
\label{Fig:RFpuro}
\end{figure}

Figure \ref{Fig:RFpuro} proposes the graphical representation of a generic RF.
The algorithm is divided into four steps numbered from 0 to 3.
In Step 0, some parts of the available data are used as training data. In Step 1, bootstrap sampling is performed over the training data, obtaining $B$ samples of the same size. In Step 2, for each bootstrap sample a tree is grown, where at each split only $m$ of the available predictors can be selected. This step results in $B$ predictions for each unit. Finally, in Step 3, the RF prediction is obtained by aggregating the $B$ predictions using Eq.\eqref{Eq:RFpred}.

\subsection{Random Forest in the spatial framework} \label{ProblemRFSpatial}

Given the flexibility of RF in modelling the relationship between the response variable and the predictors (also in the case of non-linearities), one wonders if it could be effectively applied in the case of spatial data. In particular, it could be useful to model the large scale component of a spatial model (as in Eq.(\ref{eq:01})). However, some limitations have already emerged from the literature on using standard RF for spatial applications.

The first criticism arises from the inability of RF to consider directly the information coming from the spatial location of points. In fact, RF cannot take advantage of the information not included in the set of predictors during the construction of a tree \citep{hengl2018random}. 
Moreover, the spatial correlation among the data can affect the mechanisms that allow to implement RF. Indeed, the re-sampling of correlated data violates the independence assumption adopted by bootstrap to create $B$ bagged datasets. As a consequence, some bagged sets can contain more spatially closer or further locations.
Furthermore, it is worth noting that the optimisation problem adopted for searching the best predictor-cutpoint combination can be represented as an ordinary least squares (OLS) problem in a regression model \citep{saha2021random}. However, it is known that solving an OLS in a context of dependence among the data can lead to sub-optimal results, i.e. the estimated coefficients may be biased and/or their standard errors may be underestimated.
Given these issues, it is of interest to conduct a systematic review to collect and study the proposals already in the literature that have attempted to apply RF in a spatial context.

%%%%%%%%%%%%%%%%%%%%%%%%%%%%%%%%
\section{Taxonomy} \label{TaxonomySec}

In this section, we propose a new taxonomy for classifying scientific documents related to the application of regression RF for spatially correlated data. We first define three main categories, named \textit{Pre-}, \textit{In-}, and \textit{Post-processing}, according to when RF is adjusted:
\begin{enumerate}
\item \textit{Pre-processing}, the spatial information is dealt with before running actually regression RF by including predictors which are somehow informative of the spatial autocorrelation existing in the data. With respect to the graphical representation of RF (see Figure \ref{Fig:RFpuro}), this category includes the strategies undertaken at Step 0 by augmenting the available data.
\item \textit{In-processing}, this refers to strategies which perform a substantial change of RF which can happen at Step 1 (bootstrap sampling) or Step 2 (building trees) of Figure \ref{Fig:RFpuro}.
\item \textit{Post-processing}, including the strategies where the spatial correlation is taken into account after running RF, by adjusting the RF predictions.
\end{enumerate}

\begin{figure}
    \centering
    \includegraphics[width=1\textwidth]{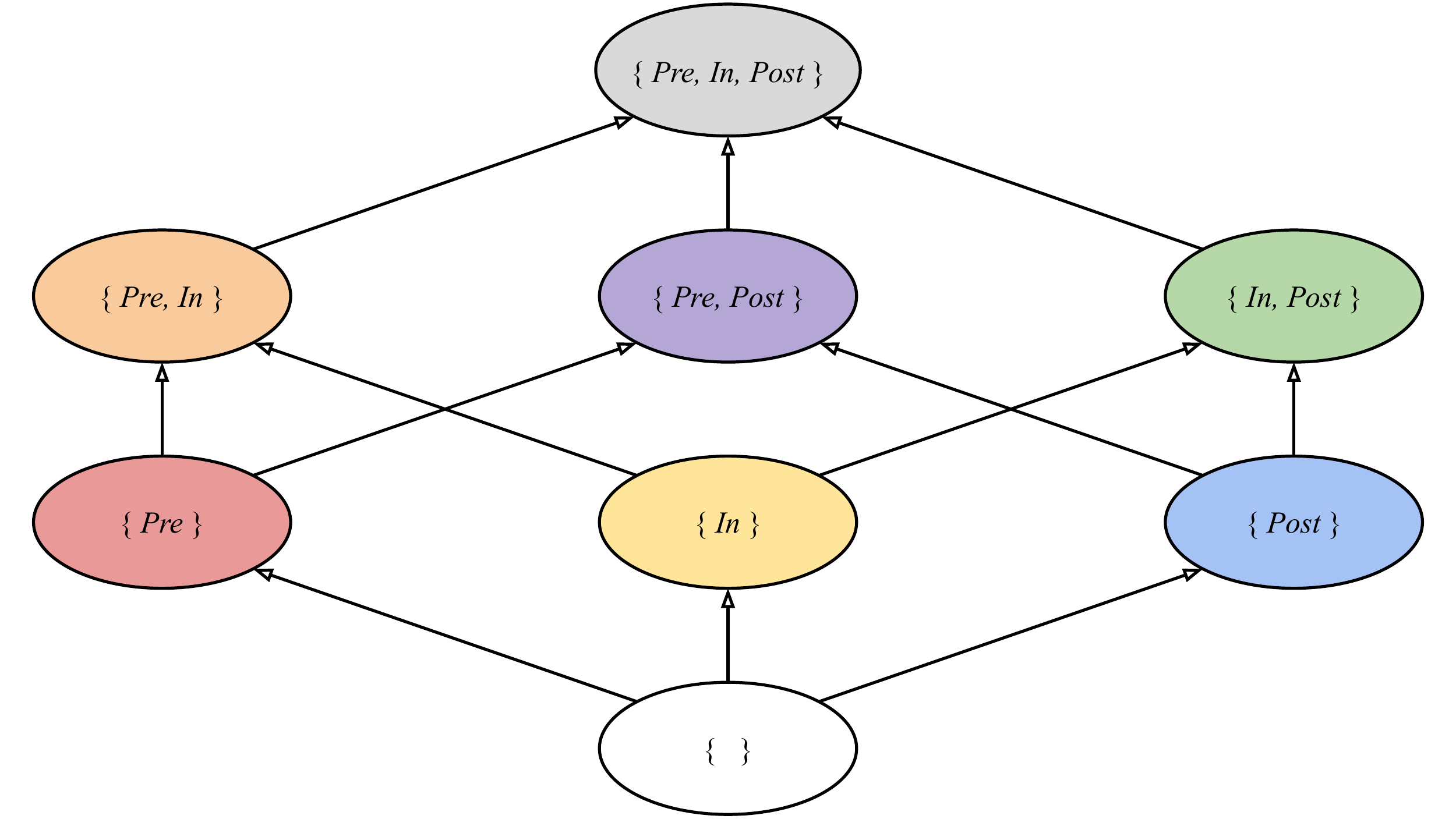}
    \caption{Graphical representation (Hasse diagram) of the power set $\mathcal{P}(S)$ with $S=\{$\text{\textit{Pre}, \textit{In}, \textit{Post}}$\}$ corresponding to the proposed taxonomy categories.}
    \label{TaxonomyTheory}
\end{figure}

However, a contribution could belong to more than one of the above-described categories giving rise to the need to define mixed categories. In this respect, we collect these three main categories in a set denoted by $S=\text{\{\textit{Pre}, \textit{In}, \textit{Post}\}}$ that gives rise to the power set $\mathcal P(S)$ (i.e. all the possible subsets of $S$) which is composed by $2^3=8$ elements (including also the empty set and $S$ itself). Basically, the power set defines the possible single and mixed categories, which are represented in Figure \ref{TaxonomyTheory} using a Hasse diagram. Note that in our case, the combinations of multiple categories are order sensitive with the intrinsic order given by the temporal sequence of the three main categories.

Starting from the bottom of Figure \ref{TaxonomyTheory}, the first subset represents the empty or null set (white), corresponding to no strategies for adjusting the RF algorithm to the spatial case. The first row reports the single categories (red, yellow and blue), while the second row refers to pairs of main categories (orange, purple, green). Finally, at the top, we find the subset with the three combined categories (grey). The arrows display the upward path from the simplest to the most complex mixed categories.

\section{State of the art: random forest in the spatial framework} \label{RFStateofArt}

\begin{figure}[t]
\centering
    \includegraphics[width=1\textwidth]{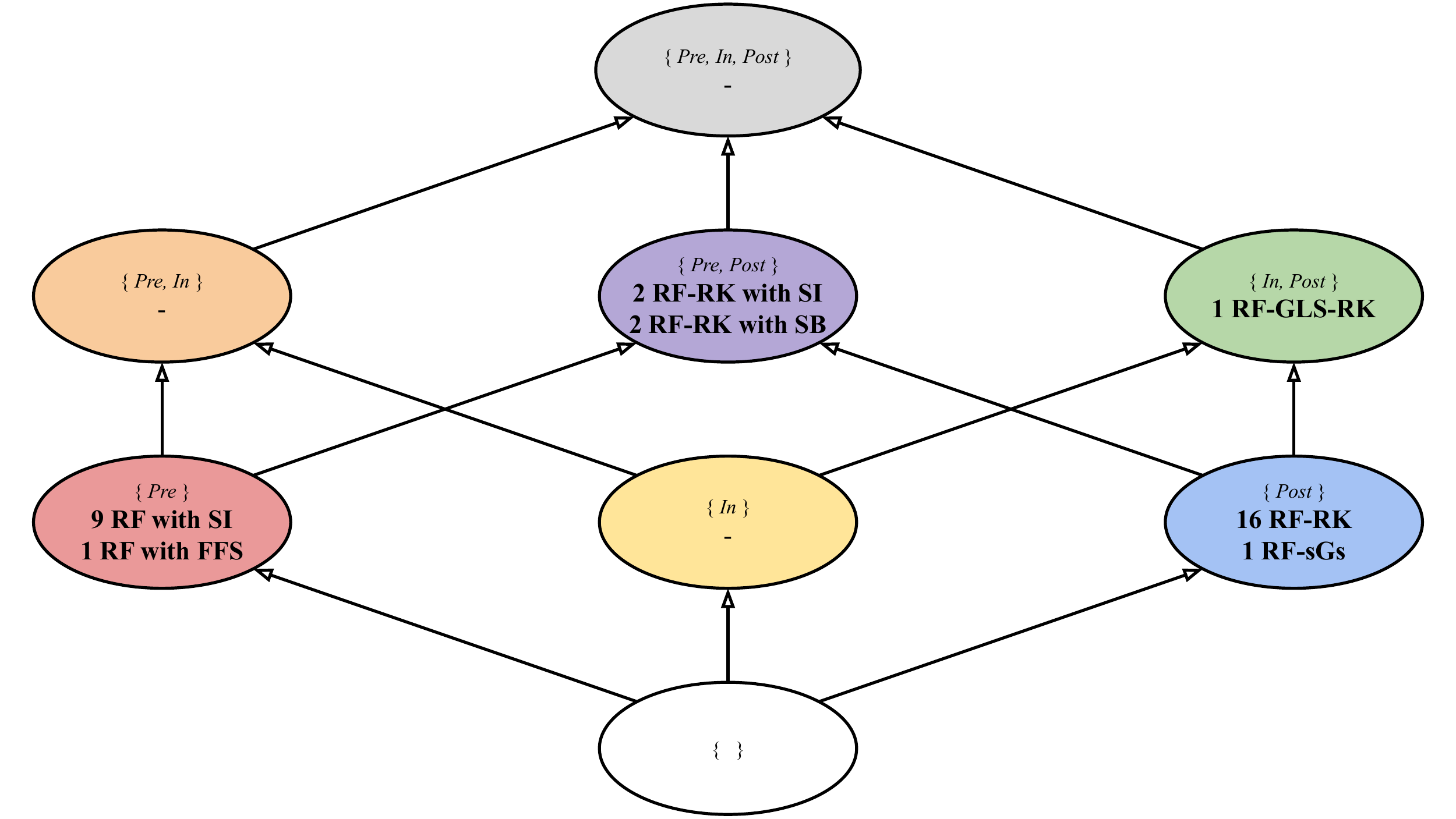}
\caption{Classification of the identified contributions according to the taxonomy. The adopted strategies are: RF with SI: Random Forest with Spatial Information; RF with FFS: Random Forest with Forward Feature Selection; RF-RK: Random Forest Residual Kriging; RF-sGs: Random Forest sequential Gaussian simulation; RF-RK with SI: Random Forest Residual Kriging with Spatial Information; RF-RK with SB: Random Forest Residual Kriging with Spatial Bootstrap; RF-GLS-RK: Random Forest based on GLS Residual Kriging.}
\label{TaxonomyFull}
\end{figure}

In this section, the 32 identified contributions (see Section \ref{PRISMA}) are classified with respect to the proposed taxonomy and then discussed. Figure \ref{TaxonomyFull} shows the Hasse diagram also including the labels of adopted strategies inside each taxonomy category and the corresponding number of documents. As expected, the null set $\{  \ \}$, representing no strategies, is empty because, as stated in Section \ref{PRISMA}, we are interested in the contributions that tried to adjust RF to spatially dependent data. Moreover, note that the three categories $\{\textit{In}\}$, $\{\text{\textit{Pre}, \textit{In}}\}$ and $\{\text{\textit{Pre}, \textit{In}, \textit{Post}}\}$ are also empty, given that none of the 32 documents falls into these groups. 
Table \ref{SystematicTable} contains a detailed list of the 32 analyzed scientific documents, including the taxonomy category, the label of adopted strategy, the reference (author(s) and year), the title and the source from which it was obtained (query search or snowballing).

In the following, we present in detail the 32 scientific documents grouped according to the taxonomy category and the adopted strategies are discussed.

\subsection{Pre-processing category} 

This category includes all the documents that deal with spatial correlation applying some strategies in the \textit{Pre-processing} phase, i.e. before running RF. In particular, one strategy is found in the literature: consider in RF new predictors as proxies of the Spatial Information (RF with SI). For this strategy, adopted by 9 out of 32 found documents, we define three subgroups named as follows: spatial Random Forest; Random Forest for Spatial Interpolation; Random forest of kriged data. All these strategies apply a data augmentation approach considering a new set of variables to grasp the spatial information. The description of such subgroup of strategies is reported here below.

\textit{Spatial Random Forest} was firstly introduced by \cite{hengl2018random} with the acronym RFsp. It includes geographical predictors based on buffer distance maps. In practice, the authors suggest adding in the predictor matrix new columns given by the pairwise geographical Euclidean distances among all sampling points (for example, the first additional column will contain the distances with respect to the first site). Thus RFsp is a standard RF algorithm, as described in Section \ref{RFSection}, where the augmented predictor matrix also contains information about the geographical proximity between observations trying to mimic spatial correlation used in Kriging. Differently from \cite{hengl2018random}, \cite{behrens2018spatial} use RFsp considering different predictors to account for geographical proximity (i.e., a set of euclidean distances from the corners and the centres of the location space in combination with absolute coordinates). They applied the proposed methodology for the prediction of soil properties. A drawback of RFsp is that the distances to be considered (and then the number of extra columns in the predictor matrix) increase with the number of spatial locations, resulting in higher computational costs \citep{WADOUX2020review}. \cite{moller2020oblique} pointed out another drawback of RFsp, especially in the context of soil properties mapping. They claimed that the use of the coordinates or the distances in the predictor set could give rise to artifacts in the final maps or difficulties in the RF interpretation. Thus they proposed using Oblique Geographic Coordinates (OGCs), i.e. coordinates which are defined along several axes tilted at oblique angles. The authors claim that by including these oblique coordinates as predictors, the decision tree algorithm makes oblique splits in the geographic space and provides a more realistic prediction surface. In the same wake, \cite{hu2022incorporating} introduced a set of proxy variables to get the spatial pattern unexplained by the predictors in the context of house price estimation: the spatial coordinates to consider the absolute geographic locations; the Moran's index together with spatial eigenvectors, generated from a contiguity-based spatial weights matrix, for capturing local spatial variability (see \citealp{griffith2006spatial} and \citealp{DRAY2006483} for more details about eigenfunction spatial analysis). A similar approach is adopted by \cite{santiago2022contrasts}, where the relation between soil properties and plants presence is addressed.
 
More recently, a second type of spatial RF has been proposed by \cite{talebi2022truly} with the acronym SRF presenting an application regarding the prediction of the concentration of minerals. This approach considers ``vectorized spatial patterns" as extra predictors that provide spatial information. In particular, the predictor matrix is augmented by including the values of the original predictors observed in a set of neighbours of the training data locations (possibly also rotated). The method was developed especially for gridded data, but it can also be applied to geostatistical data when they are rasterized.

\textit{Random Forest for Spatial Interpolation}, known as RFSI, consists in augmenting the predictor matrix, including the value of the response variable observed in neighbouring locations and their distances to the prediction locations. It was proposed by \cite{sekulic2020random} for precipitation and temperature prediction.
Based on the same intuition, a Quantile Regression Forest Spatial Interpolation (QRFI) was proposed by \cite{cordoba2021random}. Both RFSI and QRFI consider the $k$-nearest response values and distances as additional predictors ($k \leq n$). The difference is that QRFI applies Quantile RF \citep{meinshausen2006quantile} instead of standard RF. In this way, it is possible to infer the conditional distribution of the response rather than the conditional mean as standard RF does.

\textit{Random forest of kriged data}

was proposed by \cite{dhara2018machine}. The authors introduce spatial information by using as response in RF the smoothed kriged values of the response variable (i.e. porosity of the soil) with the idea that spatial correlation is taken into account once Ordinary Kriging (OK, see Eq.~\eqref{eq:01}) predictions on a grid are obtained. After OK, they perform standard RF on grid values (so modeling the relationship between the response and the predictors after OK).

\begin{table}[!htbp]
\resizebox{\textwidth}{!}{%
\centering
\input{Systematic0307}}
\caption[Caption for LOF]{List of the 32 documents identified through the PRISMA methodology and categorised by the proposed taxonomy (see Figure \ref{TaxonomyFull}). \\
Adopted strategy: RF with SI: Random Forest with Spatial Information; RF with FFS: Random Forest with Forward Feature Selection; RF-RK: Random Forest Residual Kriging; RF-sGs: Random Forest sequential Gaussian simulation; RF-RK with SI: Random Forest Residual Kriging with Spatial Information; RF-RK with SB: Random Forest Residual Kriging with Spatial Bootstrap; RF-GLS-RK: Random Forest based on GLS Residual Kriging. \\
From: Q: query; S: backward snowballing.
}
\label{SystematicTable}
\end{table}

Another strategy in the \textit{Pre-processing} phase was found in \cite{meyer2019importance} and  consists in performing variable selection before fitting RF, named RF with FFS. In particular, the authors highlight that highly spatially autocorrelated predictors can lead to considerable overfitting that can be avoided by applying a proper variable selection. 
Indeed, in modeling Leaf Area Index, they consider the spatial Forward Feature Selection algorithm (FFS) introduced in \cite{meyer2018improving}. FFS follows the ratio of a Forward Step Selection but fitting RF iteratively and evaluating performances at each step by means of a spatial Cross-Validation scheme (where the data are split into folds not randomly but according to spatial locations, with neighbouring observations expected to be in the same fold).

\subsection{Post-processing category} \label{sec:postprocessing}
This category is the most common among the PRISMA output, with 17 out of 32 scientific contributions. It includes the documents that propose to deal with spatial autocorrelation by \textit{Post-processing} the RF output. In particular, two main strategies are found: apply Kriging on the RF residuals, i.e. Random Forest Residual Kriging (RF-RK); Random Forest followed by a sequential Gaussian simulation on the residuals (RF-sGs).

\textit{Random Forest Residual Kriging}, RF-RK, is a hybrid strategy that combines the RF algorithm with the Kriging model exploiting the advantages of the two methods. On the one hand, with the RF algorithm, it is possible to handle non-linearities and a high dimensional predictor space; on the other hand, it takes explicitly into account the spatial correlation through Kriging. 
Formally, the Kriging model can be specified as follows \citep{cressie1993statistics}:
\begin{equation}\label{eq:01}
    y(s) \\ = \\ \mu(s) \\ + \\ \omega(s),
\end{equation}
where $\mu(s)$ represents the large scale component, also called \textit{trend} or \textit{drift}, while $\omega(s)$ is a zero mean spatially correlated process, for which the second order stationarity and isotropy are assumed, representing the small scale variability. 
A Kriging model is said Ordinary (OK) when $\mu(s)=\mu$ that is an unknown constant to be estimated. Instead, when the large scale term is a function of some predictors $X$, the model is known in the literature as Regression Kriging model.
RF-RK is a two-stage approach, one stage for the large scale component and one for the small scale component. Indeed, first the trend of the spatial model in Eq.\eqref{eq:01} is estimated by using RF, then 
an OK is fitted over the RF residuals giving rise to the so-called Residual Kriging (RK).
The final prediction for the response variable at an unobserved location $s_0$ will be given by 
\begin{equation}\label{FormulaRF-RK}
    \hat{y}(s_0)=\hat{\mu}_{RF}(X_0)+\hat{\omega}(s_0),
\end{equation}
where $\hat{\mu}_{RF}(X_0)$ is the fitted large scale using RF algorithm, and $\hat{\omega}(s_0)$ the kriged residual obtained using OK model.

RF-RK is adopted by \cite{fox2020comparing} for macroinvertebrate multimetric index mapping as an alternative to OK and standard RF. A by-product of this paper is the construction of prediction intervals, a procedure which is not straightforward in RF compared to standard statistical spatial models. A slight variation in RF-RK consists in using residuals obtained not from the standard RF algorithm, but from one of its variants, as for example QRF. This approach is adopted by \cite{vaysse2017using} for soil properties mapping in comparison with the standard RF-RK. Interestingly, the authors use QRF to build prediction intervals, whose length represents a measurement of prediction uncertainty. QRF has also been adopted in \cite{cordoba2021spatially} for the prediction of potential profits of fields and land value.
Another alternative version of RF-RK is proposed by \cite{szatmari2021estimating} for modeling soil organic carbon (SOC) stock: being the data available for two years, they propose to use coKriging model \citep{cressie1993statistics}, instead of the OK one, for jointly modelling the spatial distribution of SOC stocks for both years.

Furthermore, \cite{fayad2016regional} applied RF-RK with the preliminary use of the Variable Selection Using Random Forest algorithm (VSURF) proposed by \cite{genuer2010variable}. This non-spatially based variable selection procedure is used to identify the most important predictors for forest canopy heights and biomass.

Other 11 literature contributions belonging to the \textit{Post-processing} category just apply RF-RK with no particular features worth mentioning. They are listed here below with a short reference to the kind of application. In the context of ecology, \cite{dos2018spatial} predicted forest basal area and volume. \cite{guo2015digital,garcia2017estimating} and \cite{da2022soil} focused on the prediction of Soil Organic Matter (SOM), \cite{smith2022spatial} and \cite{makungwe2021performance} analysed soil pH and soil nitrogen, \cite{ahmed2017assessing} tried to predict SOC and other soil fractions, while \cite{hengl2015mapping} and \cite{mammadov2021estimation} considered different soil properties such as SOC or the presence of single soil elements. Linked to environment and pollution, we mention the paper by \cite{liu2018improve} regarding particulate matter concentrations. Finally, \cite{paccioretti2021statistical} dealt with yields and profits in the agricultural sector.

\textit{Random Forest sequential Gaussian simulation} is another example of a hybrid strategy represented by the combination of RF algorithm and conditional sequential Gaussian simulation (sGs), proposed by \cite{koch2019modeling} for studying redox depth in the context of water resources management. Similarly to RF-RK, the prediction is obtained as a sum of two elements: the large scale, estimated by the use of RF algorithm, and the small scale, obtained in this case by sGs (originally proposed by \citealp{PEBESMA1998}) instead of OK model. In particular, with sGs it is possible to simulate from the residual distribution, conditional on the observations and the estimated variogram, and the variability of the sGs realizations can be used as an uncertainty measure.

\subsection{Mixed-processing categories}
Inside the mixed-processing categories, we classify the scientific contributions that combine more than one category from the taxonomy. In particular, from the literature review, we find five documents which are reported in the $\{\textit{Pre}, \textit{Post}\}$ (purple) and $\{\textit{In}, \textit{Post}\}$ (green) subsets of Figure \ref{TaxonomyFull}.

\cite{kurina2019spatial} and \cite{li2011can} opted for a \textit{Pre-Post-processing} mixed category for the spatial prediction of glyphosate sorption coefficient and of the soil pH, respectively. This consists simply in including spatial variables (as coordinates or environmental spatial predictors) in the set of predictors when applying RF-RK (i.e., RF-RK with SI). We find two other contributions adopting a \textit{Pre-Post} category: \cite{viscarra2014mapping} for rock gamma radiation and \cite{szatmari2019comparison} for SOC stock change. Their main approach is the RF-RK described in Section \ref{sec:postprocessing}. The novelty consists in replicating the RF-RK algorithm a given number of times, say $T$, in order to obtain confidence intervals as a measure of predictions uncertainty. In particular, they adopt bootstrap for spatially dependent data to create $T$ training datasets, thus accounting for the spatial correlation in the data. The $T$ predictions they get for each location, by running RF-RK $T$ times, represent a bootstrap distribution of the predictions from which it is possible to compute any summary statistics. Finally, to quantify the overall uncertainty of RF-RK they calculate the prediction variance as the sum of the bootstrap variance and the variance from the kriged residuals.

The most interesting strategy, from a statistical point of view, is the one by \cite{saha2021random} in the \textit{In-Post-processing} category.
The proposed solution modifies the RF algorithm internally affecting Step 2 (see Figure \ref{Fig:RFpuro}).
The authors started recalling the regression tree optimization problem can be written as an OLS problem with membership in current leaf nodes forming the OLS design matrix, moving from a local to a global optimization problem. In case of spatially correlated data OLS assumptions are violated and then \cite{saha2021random} propose to include the spatial correlation structure by moving to a Generalised Least Squares (GLS) criterion (as it is usually done in linear models). The new algorithm is called Dependency Adjusted Regression Tree (DART) and is used for fitting each tree in a Random Forest. The RF algorithm which aggregates $B$ trees, built using DART, is named RF-GLS and is implemented in the R package \texttt{RandomForestsGLS} described in \cite{saha2022randomforestsgls}.

Moreover, \cite{saha2021random} suggest fitting an OK model to the residuals obtained from the RF-GLS (i.e., RF-GLS-RK).
In more detail, the estimation procedure starts with performing a standard RF algorithm and computing the corresponding residuals. Then, before moving to the GLS step, the covariance matrix (based on a parametric spatial model as the Mat\`ern function) is estimated by fitting a zero mean Gaussian process to RF residuals using the Maximum Likelihood approach (or using Nearest Neighbor Gaussian Process for large datasets). Subsequently, an RF-GLS algorithm is performed with this estimated covariance matrix and finally OK is performed on the new residuals for prediction purposes. 

\section{Conclusions} \label{Conclusions}

In this work, we propose a new taxonomy for classifying scientific contributions focused on a specific research topic: the use of the RF algorithm for point-referenced data characterized by spatial correlation in regression problems. This is a real and crucial topic: on the one hand, RF is gaining increasing popularity in the context of spatial data, also thanks to its flexibility in modeling complex input-output relationships and a high dimensional predictor space; on the other hand, not taking into account the spatial dependence when running RF can negatively affect the predictive performance.

The taxonomy we propose is based on three main categories (\textit{Pre}, \textit{In}, \textit{Post-processing}) and some combinations of them. We applied the taxonomy classification to 32 scientific documents obtained from the systematic literature review based on the PRISMA approach. The proposed taxonomy can be used for classifying any future contribution on the same topic, or any topic where a ML algorithm has a central role (also in classification problems, obviously).

The identification of relevant contributions starts with the selection of some keywords. For this work, we decided to search scientific documents regarding random forest for spatially dependent or spatially correlated data which are specific to the geostatistics jargon. It would have been possible to include other more general keywords (e.g. spatial heterogeneity) but this would have led to a broader search and a larger number of non-relevant contributions.

Interestingly, only one document, out of 32 classified contributions, was published in a statistical journal: it is the paper by \cite{saha2021random} appeared in 2021 in the Journal of the American Statistical Association. In our opinion, Saha and colleagues contributed to the state-of-the-art with the most substantial methodological change of the standard RF algorithm. The remaining documents are published in journals with a more applicative perspective and dealing with real case studies pertaining to different fields essentially in environmental sciences, e.g. soil science, forestry, and ecology. The most adopted strategy is the RF-RK which combines the standard RF algorithm, for the large-scale estimation, with the most classical geostatistical model, i.e. Ordinary Kriging, for retrieving the spatial correlation existing in the RF residuals for prediction purposes.

One element to be taken into account when comparing the various proposals is the computational complexity and cost. For example, the strategy proposed by \cite{saha2021random} requires different estimation steps, including the computation of the inverse of the covariance matrix. Other strategies require the calculation of buffer distances and/or the inclusion of numerous predictors, such as in \cite{hengl2018random} and \cite{behrens2018spatial}. 

Note that in a comparative perspective, only a few of the identified papers (\citealt{hengl2018random,sekulic2020random,talebi2022truly}) actually discuss the computational costs or even the computational efficiency of the adopted methods.

In conclusion, it can be said that the debate for finding a solution for applying the RF algorithm to a spatial context is still open. This is a very active research topic that is expected to return new methodological and applicative contributions in the recent future. This review can be extended to classification problems and to spatio-temporal data. In this respect, the proposed taxonomy could be directly applied. Moreover, for a deeper comparison and discussion of all the existing strategies, a wide simulation study would be desirable. However, this is beyond the purpose of this review and could be considered as future research.

%%%%%%%%%%%%%%%%%%%%%%%%%%%%

\section*{Acknowledgments}
This work was partially funded by Fondazione Cariplo under the grant 2020-4066 ``AgrImOnIA: the impact of agriculture on air quality and the COVID-19 pandemic'' from the ``Data Science for science and society'' program.

\bibliographystyle{agsm}
\bibliography{References.bib}

\end{document}

%% file: Systematic0307.tex
   \begin{tabular}{p{1.4cm}p{1.9cm}p{2.8cm}p{8.7cm}p{0.6cm}}
   %\begin{tabular}{p{0.09\textwidth}p{0.12\textwidth}p{0.19\textwidth}p{0.55\textwidth}p{0.05\textwidth}}
\toprule
\textbf{Taxonomy} & \textbf{Strategy} & \textbf{Reference} & \textbf{Title}  & \textbf{From} \\
\textbf{Category} &  &  &  & \\
\midrule
 \textit{Pre} & RF with SI & \cite{behrens2018spatial} & \textit{Spatial modelling with Euclidean distance fields and machine learning} & \hfil Q \\
 &  & \cite{dhara2018machine} & \textit{Machine learning based methods for estimation and stochastic simulation} & \hfil Q \\
 &  & \cite{hengl2018random} & \textit{Random forest as a generic framework for predictive modeling of spatial and spatio-temporal variables} & \hfil S \\
 &  & \cite{moller2020oblique} & \textit{Oblique geographic coordinates as covariates for digital soil mapping} & \hfil S \\
 &  & \cite{sekulic2020random} & \textit{Random forest spatial interpolation} & \hfil S \\
 &  & \cite{cordoba2021random} & \textit{A random forest-based algorithm for data-intensive spatial interpolation in crop yield mapping} & \hfil S \\
 &  & \cite{hu2022incorporating} & \textit{Incorporating spatial autocorrelation into house sale price prediction using random forest model} & \hfil Q \\
 &  & \cite{santiago2022contrasts} & \textit{Contrasts among cationic phytochemical landscapes in the southern United States} & \hfil Q \\
 &  & \cite{talebi2022truly} & \textit{A truly spatial random forests algorithm for geoscience data analysis and modelling} & \hfil Q \\
 \cmidrule{2-5}
 & RF with FFS & \cite{meyer2019importance} & \textit{Importance of spatial predictor variable selection in machine learning applications} & \hfil S \\
 \hline
 %
%\textit{In} & SCV & \cite{meyer2019importance} & \textit{Importance of spatial predictor variable selection in machine learning applications} & \hfil S \\
 %&  & \cite{deppner2022accounting} & \textit{Accounting for spatial autocorrelation in algorithm-driven hedonic models: a spatial cross-validation approach} & \hfil Q \\ \hline
%
 \textit{Post} & RF-RK & \cite{guo2015digital} & \textit{Digital mapping of soil organic matter for rubber plantation at regional scale: an application of random forest plus residuals kriging approach} & \hfil S \\
 &  & \cite{hengl2015mapping} & \textit{Mapping soil properties of Africa at 250 m resolution: random forests significantly improve current predictions} & \hfil S \\
 &  & \cite{fayad2016regional} & \textit{Regional scale rain-forest height mapping using regression-kriging of spaceborne and airborne LiDAR data: application on French Guiana} & \hfil S \\
 &  & \cite{ahmed2017assessing} & \textit{Assessing soil carbon vulnerability in the Western USA by geospatial modeling of pyrogenic and particulate carbon stocks} & \hfil S \\
 &  & \cite{garcia2017estimating} & \textit{Estimating soil organic matter using interpolation methods with a electromagnetic induction sensor and topographic parameters: a case study in a humid region} & \hfil Q \\
 &  & \cite{vaysse2017using} & \textit{Using quantile regression forest to estimate uncertainty of digital soil mapping products} & \hfil S \\
 &  & \cite{dos2018spatial} & \textit{Spatial prediction of basal area and volume in Eucalyptus stands using Landsat TM data: an assessment of prediction methods}& \hfil Q\\
  &  & \cite{liu2018improve} & \textit{Improve ground-level PM2.5 concentration mapping using a random forests-based geostatistical approach} & \hfil Q \\
  &  & \cite{fox2020comparing} & \textit{Comparing spatial regression to random forests for large environmental data sets} & \hfil S \\
 &  & \cite{cordoba2021spatially} & \textit{A spatially based quantile regression forest model for mapping rural land values} & \hfil Q \\
 &  & \cite{paccioretti2021statistical} & \textit{Statistical models of yield in on-farm precision experimentation} & \hfil Q \\
 &  & \cite{makungwe2021performance} & \textit{Performance of linear mixed models and random forests for spatial prediction of soil Ph} & \hfil Q \\
 &  & \cite{mammadov2021estimation} & \textit{Estimation and mapping of surface soil properties in the Caucasus Mountains, Azerbaijan using high-resolution remote sensing data} & \hfil Q \\
 &  & \cite{szatmari2021estimating} & \textit{Estimating soil organic carbon stock change at multiple scales using machine learning and multivariate geostatistics} & \hfil Q \\
 &  & \cite{da2022soil} & \textit{Soil organic matter and clay predictions by laboratory spectroscopy: data spatial correlation} & \hfil Q \\
 &  & \cite{smith2022spatial} & \textit{Spatial variability and uncertainty of soil nitrogen across the conterminous United States at different depths} & \hfil Q \\
 \cmidrule{2-5}
 & RF-sGs & \cite{koch2019modeling} & \textit{Modeling depth of the redox interface at high resolution at national scale using random forest and residual Gaussian simulation} & \hfil Q \\ \hline
 
\textit{In-Post} & RF-GLS-RK & \cite{saha2021random} & \textit{Random forests for spatially dependent data} & \hfil Q \\
\hline
\textit{Pre-Post} & RF-RK with SI & \cite{li2011can} & \textit{Can we improve the spatial predictions of seabed sediments? A case study of spatial interpolation} & \hfil S \\
 & & \cite{kurina2019spatial} & \textit{Spatial predictive modelling essential to assess the environmental impacts of herbicides} & \hfil Q \\
 \cmidrule{2-5}
 &RF-RK with SB& \cite{viscarra2014mapping} & \textit{Mapping gamma radiation and its uncertainty from weathering products in a Tasmanian landscape with a proximal sensor and random forest kriging} & \hfil S \\
 & & \cite{szatmari2019comparison} & \textit{Comparison of various uncertainty modelling approaches based on geostatistics and machine learning algorithms} & \hfil S \\
\bottomrule
\end{tabular}